\documentclass[final]{cvpr}

\usepackage{listings}
\usepackage{times}
\usepackage{epsfig}
\usepackage{graphicx}
\usepackage{amsmath}
\usepackage{amssymb}
\usepackage{listings}

\usepackage[pagebackref=true,breaklinks=true,colorlinks,bookmarks=false]{hyperref}

\def\cvprPaperID{****} % *** Enter the CVPR Paper ID here
\def\confYear{CVPR 2021}

\begin{document}

%%%%%%%%% TITLE
\title{AugLy: Data Augmentations for Robustness}

\author{Zo{\"e} Papakipos and Joanna Bitton \\
Meta AI \\
% Institutiona1 address\\
{\tt\small \{zoep, jbitton\}@fb.com}\\\\
% For a paper whose authors are all at the same institution,
% omit the following lines up until the closing ``}''.
% Additional authors and addresses can be added with ``\and'',
% just like the second author.
}
% }
% }

\makeatletter
\let\@oldmaketitle\@maketitle% Store \@maketitle
\renewcommand{\@maketitle}{\@oldmaketitle% Update \@maketitle to insert...
  \centering\includegraphics[width=200px]
    {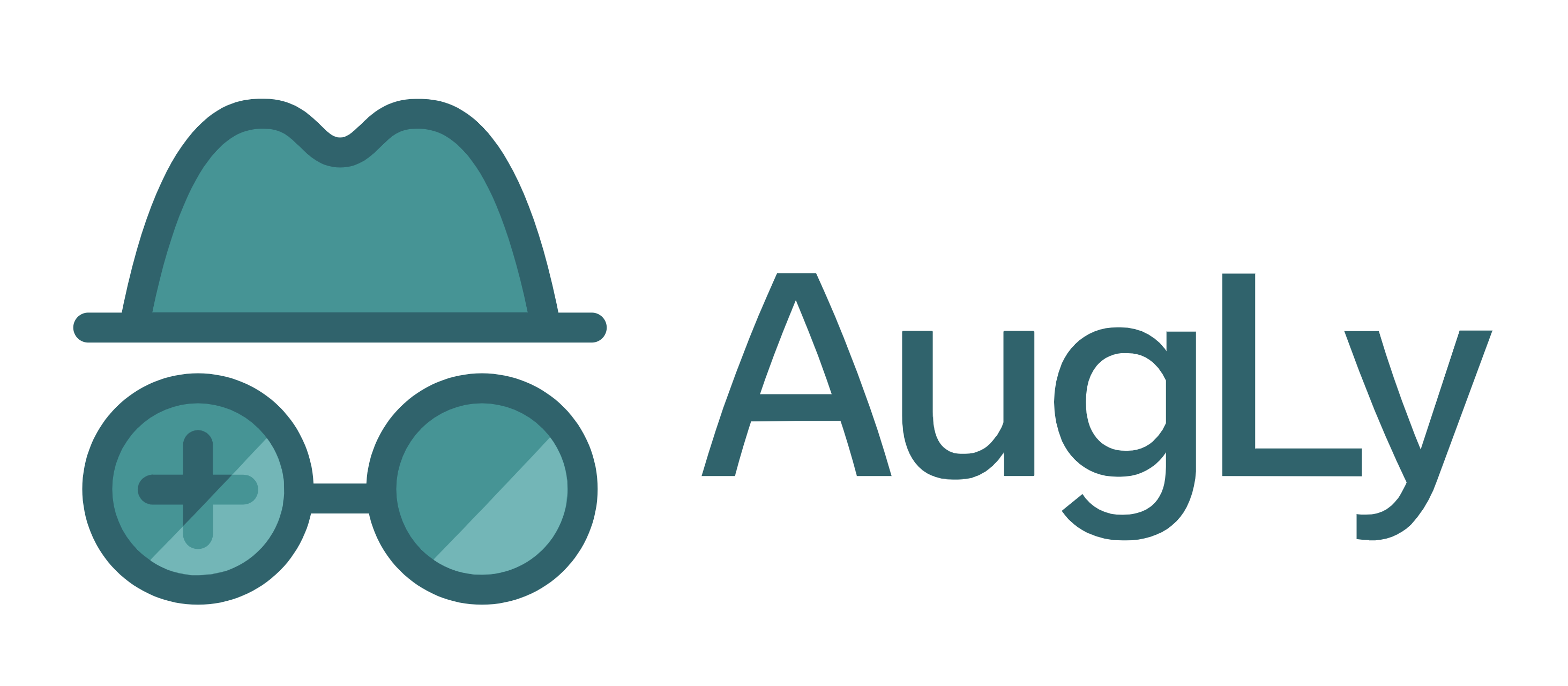}\bigskip\bigskip}% ... an image
\makeatother

\maketitle

%%%%%%%%% ABSTRACT
\begin{abstract}
   We introduce AugLy, a data augmentation library with a focus on adversarial robustness. AugLy provides a wide array of augmentations for multiple modalities (audio, image, text, \& video). These augmentations were inspired by those that real users perform on social media platforms, some of which were not already supported by existing data augmentation libraries. AugLy can be used for any purpose where data augmentations are useful, but it is particularly well-suited for evaluating robustness and systematically generating adversarial attacks. In this paper we present how AugLy works, benchmark it compared against existing libraries, and use it to evaluate the robustness of various state-of-the-art models to showcase AugLy's utility. The AugLy repository can be found at \url{https://github.com/facebookresearch/AugLy}
\end{abstract}

%%%%%%%%% BODY TEXT
\section{Introduction}

Data augmentations are a key component in the computer vision model development life cycle\cite{data_augmentations_survey}, and are also becoming increasingly prevalent in other domains\cite{nlp_data_augmentations}. They are commonly used to increase the size of datasets and prevent overfitting by performing perturbations on the input data. In addition to the classical use cases, data augmentations can also be used to evaluate the robustness of trained models to perturbations not seen at train time\cite{robustness_perturbations}\cite{robustness_ood}.

For instance, to preserve a sense of data provenance, being robust to data manipulations is critical. Content online is often manipulated and reshared, for example when users screenshot \& share a post, or overlay text or images on top of an image to make a meme. It is therefore non-trivial to be able to detect that two pieces of media are near-duplicates \cite{copy_detection}. Additionally, adversaries may try to intentionally pass in obfuscated data to a model to evade detection.

\begin{figure}[h!]
\centering
\includegraphics[width=80mm]{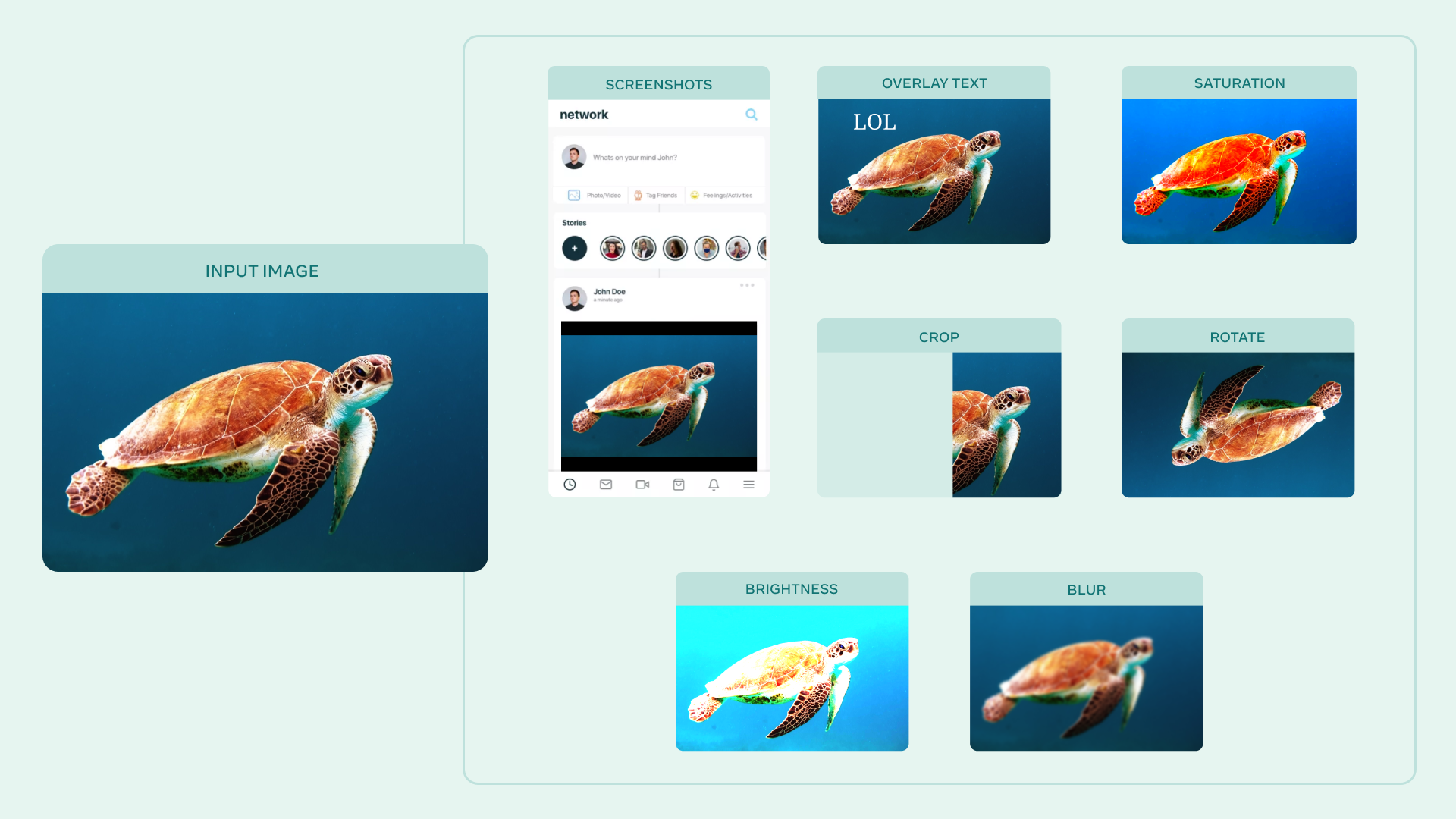}
\caption{Examples of a few AugLy image augmentations}
\label{fig:img-augs}
\end{figure}

The classical set of data augmentations used during model development does not completely mimic the way individuals online organically perturb data. Most classical augmentation libraries focus on simple transformations such as mirroring, rotating, cropping, brightness changes, etc. While these kinds of augmentations do naturally occur online, others such as overlaying text and emojis, social media screenshots, etc. are also prevalent. In addition, multimodal data processing and learning is becoming increasingly important as many real-world use cases involve multiple types of data, such as text \& images or audio \& video, and it can be useful to augment data of multiple modalities under one unified library \& API.

AugLy is built with robustness and the vast landscape of organic data augmentations seen online in mind, and to our knowledge is the first multimodal data augmentation library. AugLy can be used to synthetically create realistic data augmentations seen online, as a tool for evaluating and increasing robustness and to augment multiple modalities at a time, and thus stands out in comparison to existing libraries. In this paper we introduce AugLy, explain how it works, its architecture, and how it compares in terms of functionality \& efficiency to existing data augmentation libraries. We also conduct a robustness evaluation on state-of-the-art image classification models throughout the years to demonstrate how AugLy can be used to identify robustness gaps in pre-trained models.

%------------------------------------------------------------------------
\section{Related Work}

Most commonly-used augmentation libraries focus on one modality and provide a fairly limited set of augmentations. A majority of libraries focus on images \cite{albumentations,torchvision,imgaug} and text \cite{nlpaug,textflint,textattack}, however audio \cite{librosa,torchaudio,wavaugment} and video \cite{moviepy,pytorchvideo,vidaug} augmentation libraries do exist as well with more limited augmentations (see Section~\ref{section:benchmarking} for in-depth comparisons between AugLy and existing libraries for each modality). Meanwhile, AugLy provides augmentations for audio, images, text and video under a unified API, and is one of few libraries\cite{textflint} that focus on evaluating robustness rather than augmenting a dataset at train time. 

Other works have conducted experiments to find sets of augmentations that when trained on improve robustness at test time, such as AugMix\cite{augmix}. Strategies like AutoAugment, on the other hand, find an ``optimal'' set of augmentations to train on in a more automated way\cite{autoaugment}.

In AI Fairness, studies assessing the robustness of models to various protected categories are common. In NLP, there are studies that augment text to assess a model's biases towards gender \cite{fairness_gender_aug,gender_bias_dialogue} and ethnicity\cite{fairness_name_aug}. AugLy provides ``fairness augmentations'' since being robust to perturbations in protected classes is an important aspect of robustness that we must evaluate to ensure that models are not amplifying biases.

%------------------------------------------------------------------------
\section{AugLy}

AugLy is a novel open-source data augmentation library which provides over 100 data augmentations across four modalities: audio, image, text, and video. The augmentations provided in AugLy are informed by the perturbations that real people on the Internet perform on data daily. This includes augmentations such as overlaying text, emojis, and screenshot transforms for image \& video and inserting punctuation or similar characters for text.

\begin{figure}[h!]
\centering
\includegraphics[width=80mm]{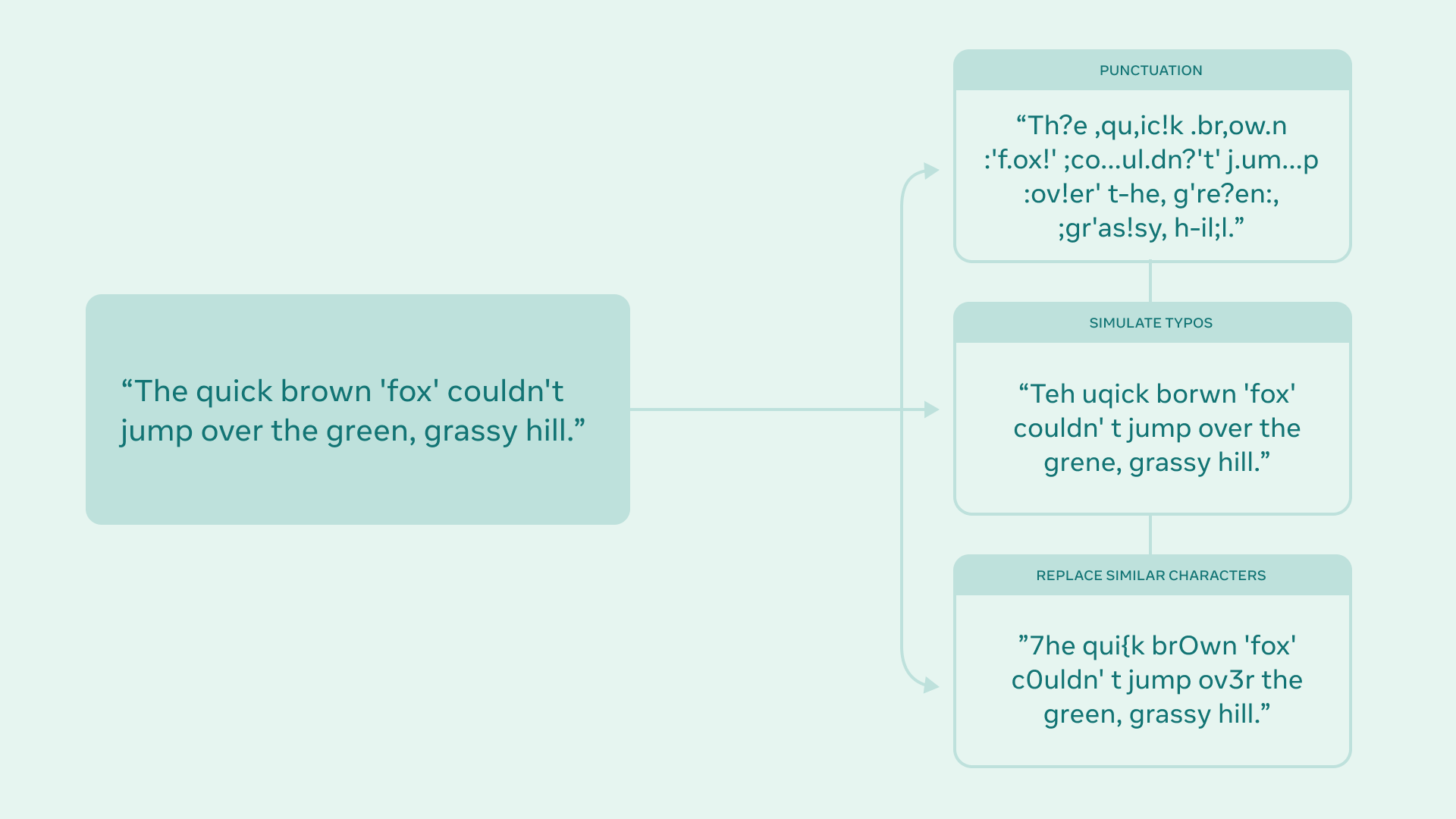}
\caption{Examples of some AugLy text augmentations}
\label{fig:txt-augs}
\end{figure}

\subsection{Library Structure}

\begin{figure}
    \centering
\begin{lstlisting}[language=Python]
import augly.image as imaugs
    
aug_img = imaugs.meme_format(
    input_img,
    caption_height=75,
    meme_bg_color=(0, 0, 0),
    text_color=(255, 255, 255),
)
\end{lstlisting}
\caption{Calling an image augmentation}
\label{fig:code-img}
\end{figure}

AugLy has four sub-libraries (audio, image, text, \& video), each corresponding to a different modality. All sub-libraries follow the same interface: we provide transforms in both function-based and class-based formats, and we provide intensity functions that compute a notion of how strong a transformation is based on the given parameters. AugLy can optionally generate metadata that provides additional context as to how the data was transformed, which is useful to perform comparisons of model performance based on the augmentation type \& intensity.

AugLy also provides operators for composing multiple augmentations together, applying augmentations with a given probability, and applying multimodal augmentations (for example augmenting both the audio \& frames in a video).

We provide many basic augmentations that are already supported in existing libraries, as well as some new transformation types that are directly informed by data perturbations observed online. For example, one of our augmentations takes an image or video and overlays it onto a social media interface to make it seem as if the image or video was screenshotted by a user on a social network. This augmentation is beneficial because individuals on the internet commonly reshare content this way, and it is important for systems to be able to identify that the content is still the same despite the added interface elements.

\begin{figure}
\centering
\begin{lstlisting}[language=Python]
import augly.video as vidaugs
import augly.audio as audaugs
    
aud_augs = audaugs.Compose(
    [
        audaugs.AddBackgroundNoise(),
        audaugs.Tempo(factor=2.0),
    ],
)
vid_augs = vidaugs.Compose(
    [
        vidaugs.Rotate(p=0.5),
        vidaugs.TimeCrop(
            offset_factor=0.2,
            duration_factor=0.4,
        ),
        vidaugs.AugmentAudio(
            audio_aug_function=aud_augs,
        ),
    ],
)
vid_augs(video_path, out_path)
\end{lstlisting}
\caption{Composing audio \& video augmentations}
\label{fig:code-audvid}
\end{figure}

\subsection{Existing Use Cases}

AugLy has already been used by several projects. SimSearchNet\cite{ssn}, an image copy detection model, was trained using AugLy augmentations. AugLy was used to evaluate the robustness of deepfake detection models in the 2019 Deepfake Detection Challenge\cite{dfdc}, ultimately influencing who were the top five winners. The dataset (DISC21) for the Image Similarity Challenge\cite{isc}, a NeurIPS 2021 competition on image copy detection, was built using AugLy as well.

%------------------------------------------------------------------------
\section{Benchmarking}
\label{section:benchmarking}

In order to show how AugLy fits into the existing ecosystem of data augmentation libraries, we compare each modality's sub-library within AugLy to a few of the most popular augmentations libraries in that respective modality. Specifically, we compare the overall focus and functionality of each library, and perform runtime benchmarking to evaluate how efficient AugLy's augmentations are. Note: the augmentations were benchmarked using AugLy v0.2.1, available on Pypi and GitHub. To see the full list of augmentations benchmarked, please review the Appendix. 

\subsection{Audio}

We chose to compare AugLy's audio augmentations to three existing and popular libraries: pydub\cite{pydub}, torchaudio\cite{torchaudio}, and audiomentations\cite{audiomentations}. See Figure \ref{fig:audio_libraries} to compare the number of distinct augmentations provided. \\

\begin{figure}
\centering
\begin{tabular}{|c|c|}
    \hline
    \textbf{Library} & \textbf{\# augmentations} \\
    \hline
    pydub & 10 \\
    \hline
    AugLy & 20 \\
    \hline
    audiomentations & 25 \\
    \hline
    torchaudio & 58 \\
    \hline
\end{tabular}
\caption{The audio augmentation libraries we chose to compare and their corresponding number of augmentations at the time of writing.}
\label{fig:audio_libraries}
\end{figure}

Each library has a slightly different focus: torchaudio and audiomentations integrate easily with pytorch (torchaudio's can also be GPU-accelerated) and are clearly intended to be used at train time to improve generalization of audio machine learning models. Pydub provides more general-purpose audio processing functionality without much emphasis on either integrating with ML training or evaluation pipelines; the number of transformation functions in Pydub is also much lower than the other three.

We benchmark each audio augmentation in AugLy, as well as some analogues that exist in the other libraries. See Figure \ref{fig:audio_benchmarking} for the runtime in seconds of each augmentation in \textbf{(1)} AugLy, \textbf{(2)} pydub, \textbf{(3)} torchaudio, \& \textbf{(4)} audiomentations. \\

\begin{figure}[h!]
\centering
\begin{tabular}{|c|c|c|c|c|}
    \hline
    \textbf{Augmentation} & \textbf{(1)} & \textbf{(2)} & \textbf{(3)} & \textbf{(4)} \\
    \hline
    PitchShift & 1.238 & & \textbf{0.372} & 0.651 \\
    \hline
    TimeStretch & 0.415 & & \textbf{0.053} & 0.121 \\
    \hline
    Reverb & 0.271 & & \textbf{0.267} & \\
    \hline
    AddBackgroundNoise & 0.048 & & & \textbf{0.019} \\
    \hline
    ChangeVolume & 0.035 & \textbf{3e-5} & 0.034 & 0.004 \\
    \hline
    HighPassFilter & 0.017 & \textbf{3e-4} & 0.017 & 0.413 \\
    \hline
    ToMono & \textbf{0.016} & & 0.022 & \\
    \hline
    Normalize & 0.015 & \textbf{4e-5} & 0.043 & 0.004 \\
    \hline
    LowPassFilter & 0.014 & \textbf{5e-4} & 0.013 & 0.163 \\
    \hline
    Clip & \textbf{0.002} & & & 0.003 \\
    \hline
    Speed & 0.002 & \textbf{6e-5} & & \\
    \hline
\end{tabular} \\
\caption{The runtime (in seconds) of audio augmentations in \textbf{(1)} AugLy, \textbf{(2)} pydub, \textbf{(3)} torchaudio, \& \textbf{(4)} audiomentations.}
\label{fig:audio_benchmarking}
\end{figure}

\subsection{Image}

We compare AugLy's image augmentations to three well-established libraries: imgaug\cite{imgaug}, torchvision\cite{torchvision}, and Albumentations\cite{albumentations}. See Figure \ref{fig:image_libraries} for a comparison of the four libraries in terms of the number of distinct augmentations provided. \\

\begin{figure}
\centering
\begin{tabular}{|c|c|}
    \hline
    \textbf{Library} & \textbf{\# augmentations} \\
    \hline
    torchvision & 28 \\
    \hline
    AugLy & 34 \\
    \hline
    Albumentations & 54 \\
    \hline
    imgaug & 179 \\
    \hline
\end{tabular}
\caption{The image augmentation libraries we chose to compare and their corresponding number of augmentations at the time of writing.}
\label{fig:image_libraries}
\end{figure}

Whereas imgaug, torchvision, and Albumentations are all geared toward providing general image augmentations to be used in computer vision training pipelines for regularization purposes, AugLy is more focused on replicating image transformations that users perform online. For example none of the other three libraries contain overlay augmentations (e.g. ``OverlayText'',  ``OverlayEmoji'', or ``OverlayOntoScreenshot''), although these are extremely common image manipulations.

This indicates a gap in existing image augmentation libraries: models are not being trained to be invariant to data manipulations that they will see in the real world. For instance, a model that detects violent or harmful content in images on any online platform needs to be invariant to the augmentations provided in AugLy; otherwise a user can bypass that model by overlaying an emoji onto the harmful image or overlaying the image onto a background.

We benchmark each AugLy image augmentation, as well as any analogues that exist in the other libraries. See Figure \ref{fig:image_benchmarking} for the runtime in seconds of each augmentation in \textbf{(1)} AugLy, \textbf{(2)} imgaug, \textbf{(3)} torchvision, \& \textbf{(4)} Albumentations. \\

\begin{figure}[h!]
\centering
\begin{tabular}{|c|c|c|c|c|}
    \hline
    \textbf{Augmentation} & \textbf{(1)} & \textbf{(2)} & \textbf{(3)} & \textbf{(4)} \\
    \hline
    PerspectiveTransform & 0.333 & 0.032 & 0.076 & \textbf{0.013} \\
    \hline
    Sharpen & 0.159 & 0.021 & 0.141 & \textbf{0.005} \\
    \hline
    ColorJitter & 0.108 & 0.038 & 0.107 & \textbf{0.015} \\
    \hline
    Blur & 0.097 & 0.013 & 0.143 & \textbf{0.005} \\
    \hline
    Saturation & 0.091 & 1.301 & 0.057 & \textbf{0.015} \\
    \hline
    Pixelization & 0.081 & \textbf{0.034} & & \\
    \hline
    Brightness & 0.078 & & 0.056 & \textbf{0.005} \\
    \hline
    Resize & 0.056 & 0.014 & 0.050 & \textbf{0.006} \\
    \hline
    EncodingQuality & 0.041 & 0.050 & & \textbf{0.002} \\
    \hline
    Contrast & 0.031 & \textbf{0.007} & 0.074 & \\
    \hline
    Rotate & 0.024 & 0.019 & \textbf{0.011} & 0.028 \\
    \hline
    Pad & 0.010 & 0.018 & \textbf{0.005} & 0.008 \\
    \hline
    ApplyLambda & 0.008 & & & \textbf{2e-5} \\
    \hline
    Grayscale & 0.005 & 0.030 & 0.002 & \textbf{0.001} \\
    \hline
    HFlip & 0.005 & 0.002 & 0.003 & \textbf{0.001} \\
    \hline
    VFlip & 0.003 & 0.001 & 0.002 & \textbf{0.001} \\
    \hline
    Crop & 0.001 & 0.008 & 6e-4 & \textbf{2e-5} \\
    \hline
\end{tabular} \\
\caption{The runtime (in seconds) of image augmentations in \textbf{(1)} AugLy, \textbf{(2)} imgaug, \textbf{(3)} torchvision, \& \textbf{(4)} Albumentations. Albumentations consistently outperforms any other library, likely due to the fact that it uses NumPy arrays as opposed to PIL. We continue to use PIL because it allows for (a) an easy integration with torchvision's \texttt{Compose()} and (b) better code readability.}
\label{fig:image_benchmarking}
\end{figure}

\subsection{Text}

We compare AugLy's text augmentations to three existing text libraries: nlpaug\cite{nlpaug}, TextAttack\cite{textattack}, \& textflint\cite{textflint}. See Figure \ref{fig:text_libraries} for a comparison of the five libraries in terms of the number of distinct augmentations provided. \\

\begin{figure}
\centering
\begin{tabular}{|c|c|}
    \hline
    \textbf{Library} & \textbf{\# augmentations} \\
    \hline
    TextAttack & 13 \\
    \hline
    AugLy & 16 \\
    \hline
    nlpaug & 16 \\
    \hline
    textflint & 55 \\
    \hline
\end{tabular}
\caption{The text augmentation libraries we chose to compare and their corresponding number of augmentations at the time of writing.}
\label{fig:text_libraries}
\end{figure}

One significant difference between AugLy and the other text augmentation libraries is the prevalence of syntactic versus semantic (i.e. character-level vs word-level) augmentations. Most augmentations in nlpaug and TextAttack are semantic (e.g. words being swapped for synonyms or antonyms), or a few simple syntactic ones (e.g. deleting/adding characters, replacing characters with nearby ones on the keyboard). AugLy provides many syntactic augmentations that are often used online in an attempt to evade detection, such as inserting punctuation, zero-width, or bidirectional characters and changing fonts.

We benchmark each AugLy text augmentation, as well as any analogues that exist in the other libraries. See Figure \ref{fig:text_benchmarking} for the runtime in seconds of each augmentation in \textbf{(1)} AugLy, \textbf{(2)} nlpaug, \textbf{(3)} TextAttack, \& \textbf{(4)} textflint. \\

\begin{figure}[h!]
\centering
\begin{tabular}{|c|c|c|c|c|}
    \hline
    \textbf{Augmentation} & \textbf{(1)} & \textbf{(2)} & \textbf{(3)} & \textbf{(4)} \\
    \hline
    SimulateTypos & 0.276 & 0.101 & 0.006 & \textbf{4e-4} \\
    \hline
    SwapGendered & & & & \\
    Words & 0.102 & & & \textbf{0.003} \\
    \hline
    Replace & & & & \\
    SimilarChars & 0.102 & 0.101 & 0.006 & \textbf{0.001} \\
    \hline
    SplitWords & 0.101 & \textbf{0.101} & & \\
    \hline
    Contractions & 0.001 & & \textbf{1e-4} & 2e-4 \\
    \hline
    ChangeCase & 4e-4 & & & \textbf{3e-4} \\
    \hline
    Insert & & & & \\
    Punctuation & & & & \\
    Chars & \textbf{1e-4} & & 0.002 & 6e-4 \\
    \hline
\end{tabular} \\
\caption{The runtime (in seconds) of text augmentations in \textbf{(1)} AugLy, \textbf{(2)} nlpaug, \textbf{(3)} TextAttack, \& \textbf{(4)} textflint.}
\label{fig:text_benchmarking}
\end{figure}

\subsection{Video}

We compare AugLy's video augmentations to three existing libraries: moviepy\cite{moviepy}, pytorchvideo\cite{pytorchvideo}, and vidaug\cite{vidaug}. See Figure \ref{fig:video_libraries} for a comparison of the four libraries in terms of the number of distinct augmentations provided. \\

\begin{figure}
\centering
\begin{tabular}{|c|c|}
    \hline
    \textbf{Library} & \textbf{\# augmentations} \\
    \hline
    pytorchvideo & 19 \\
    \hline
    moviepy & 30 \\
    \hline
    vidaug & 40 \\
    \hline
    AugLy & 43 \\
    \hline
\end{tabular}
\caption{The video augmentation libraries we chose to compare and their corresponding number of augmentations at the time of writing.}
\label{fig:video_libraries}
\end{figure}

Most existing video augmentations either focus on manipulating the spatial dimension \textit{or} the temporal dimension, as opposed to both. For instance, many individuals apply spatial image augmentations frame by frame onto videos; pytorchvideo provides one such API to do this using the torchvision transforms. Although spatial augmentations are effective, applying temporal augmentations in tandem has been shown to improve performance\cite{videomix}.

Moviepy is more of a general video processing and editing library, but it provides both spatial and temporal manipulations such as changing the speed of the video, trimming, and spatial cropping. vidaug provides similar spatial and temporal augmentations. However, none of these existing libraries provide the option to augment the audio or to perform overlay augmentations which AugLy does provide. AugLy provides a wide array of spatiotemporal augmentations which are common online such as temporally splicing one video into another, simulating a screenshot reshare, and overlaying one video onto another. AugLy is also unique in its multimodal integration, meaning a video's audio can be transformed then recombined with the video in conjunction with other augmentations).

We benchmark each AugLy video augmentation, as well as the analogues that exist in the other libraries. See Figure \ref{fig:video_benchmarking} for the runtime in seconds of each augmentation in \textbf{(1)} AugLy, \textbf{(2)} moviepy, \textbf{(3)} pytorchvideo, \& \textbf{(4)} vidaug. \\

\begin{figure}[h!]
\centering
\begin{tabular}{|c|c|c|c|c|}
    \hline
    \textbf{Augmentation} & \textbf{(1)} & \textbf{(2)} & \textbf{(3)} & \textbf{(4)} \\
    \hline
    Loop & 2.015 & \textbf{2e-4} & & \\
    \hline
    Shift & 0.773 & & & \textbf{0.016} \\
    \hline
    Pixelization & \textbf{0.662} & & & 1.996 \\
    \hline
    AugmentAudio & 0.625 & \textbf{0.001} & & \\
    \hline
    Pad & 0.400 & \textbf{0.018} & & \\
    \hline
    TimeCrop & 0.395 & & & \textbf{1e-5} \\
    \hline
    Crop & 0.352 & 9e-5 & & \textbf{2e-5} \\
    \hline
    Rotate & 0.336 & \textbf{1e-4} & 0.202 & 0.275 \\
    \hline
    Blur & 0.307 & & \textbf{0.140} & 0.179 \\
    \hline
    VFlip & 0.297 & 9e-5 & 0.151 & \textbf{2e-5} \\
    \hline
    AddNoise & 0.297 & & & \textbf{0.036} \\
    \hline
    Resize & 0.289 & \textbf{0.015} & & \\
    \hline
    ChangeVideo & & & & \\
    Speed & 0.269 & 1e-4 & & \textbf{1e-4} \\
    \hline
    HFlip & 0.269 & 1e-4 & 0.152 & \textbf{2e-5} \\
    \hline
    Grayscale & 0.266 & \textbf{0.047} & 0.081 & \\
    \hline
    ColorJitter & 0.262 & \textbf{0.035} & 0.077 & \\
    \hline
    Brightness & 0.258 & & \textbf{0.050} & \\
    \hline
\end{tabular}
\caption{The runtime (in seconds) of video augmentations in \textbf{(1)} AugLy, \textbf{(2)} moviepy, \textbf{(3)} pytorchvideo, \& \textbf{(4)} vidaug. Although other libraries are faster, we continue to use the FFMPEG CLI because we want to be able to process large videos effectively and conserve memory, instead of storing and passing videos in memory as (3) and (4) do.}
\label{fig:video_benchmarking}
\end{figure}

%------------------------------------------------------------------------
\section{Robustness Evaluation}

To demonstrate how AugLy can be used to evaluate robustness, we evaluated a few ImageNet models throughout the years on AugLy augmentations. We were interested to see how robustness has evolved as models' accuracy has improved, as well as understanding which augmentations the models were particularly vulnerable to. We chose three models to evaluate: VGG16\cite{vgg}, Resnet152\cite{resnet}, and Efficientnet-L2 (Noisy Student)\cite{efficientnet}.

We evaluated the aforementioned models on the ImageNet validation set, which is commonly used since the test set is not available for download. However, to avoid any potential bias due to overfitting, we evaluated on an additional dataset, ImageNet V2. ``ImageNet V2''\cite{imagenet_v2} was put together by researchers with the intention to be a held-out test set for ImageNet that can be evaluated on with no risk of overfitting.

\begin{figure}[h!]
\centering
\begin{tabular}[b]{|cc|}
\hline
\multicolumn{2}{|c|}{\textbf{Original image}} \\
\multicolumn{2}{|c|}{\includegraphics[width=0.4\linewidth]{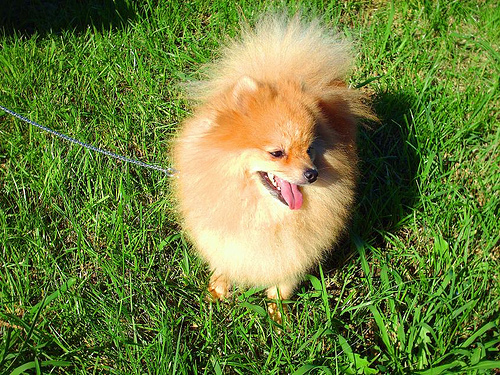}} \\
\hline
\multicolumn{2}{|c|}{\textbf{Spatial}} \\
\includegraphics[width=0.4\linewidth]{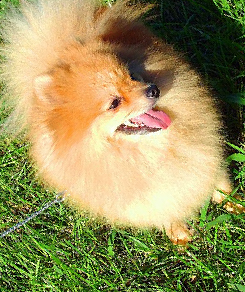} & \includegraphics[width=0.4\linewidth]{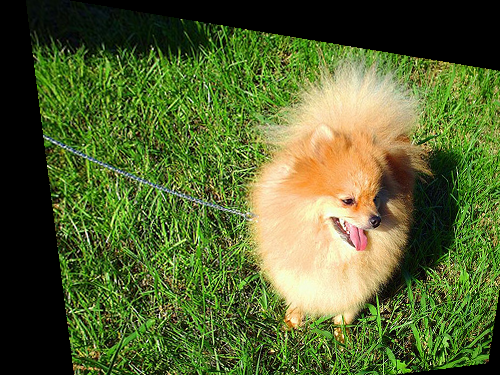} \\
rotate & perspective\_transform \\
\hline
\multicolumn{2}{|c|}{\textbf{Color}} \\
\includegraphics[width=0.4\linewidth]{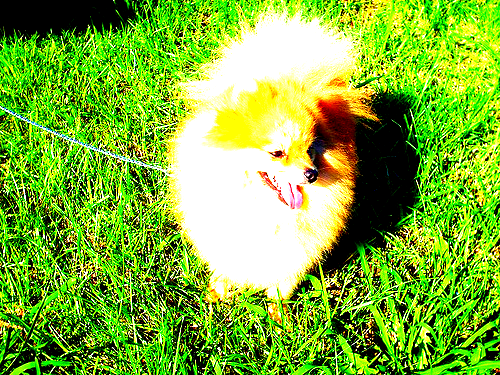} & \includegraphics[width=0.4\linewidth]{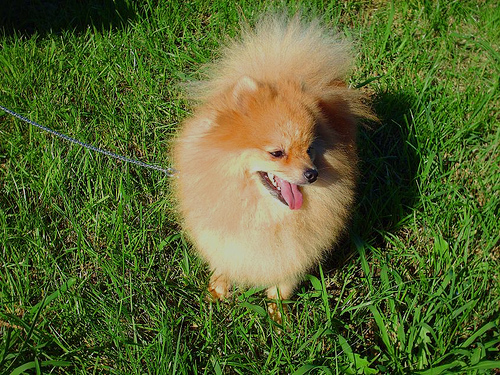} \\
color\_jitter & brightness \\
\hline
\multicolumn{2}{|c|}{\textbf{Overlay}} \\
\includegraphics[width=0.4\linewidth]{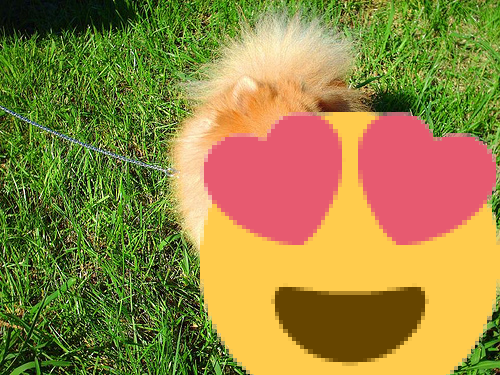} & \includegraphics[width=0.4\linewidth]{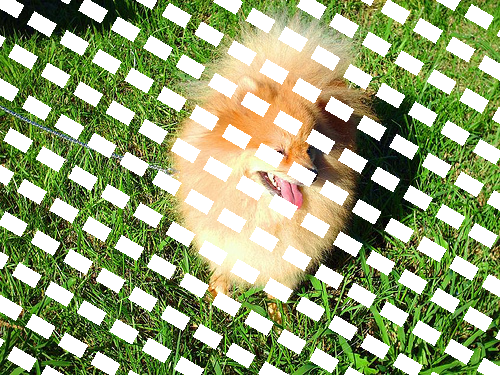} \\
overlay\_emoji & overlay\_stripes \\
\hline
\multicolumn{2}{|c|}{\textbf{Pixel-level}} \\
\includegraphics[width=0.4\linewidth]{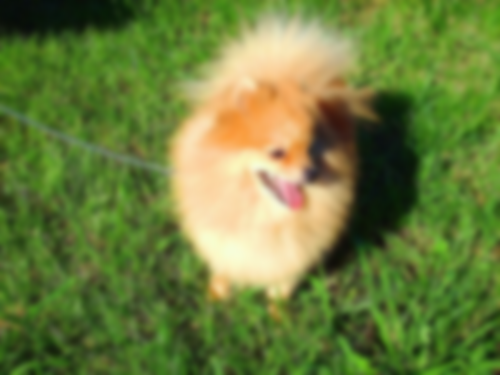} & \includegraphics[width=0.4\linewidth]{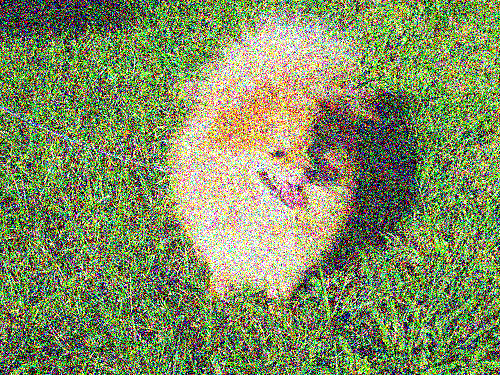} \\
blur & random\_noise \\
\hline
\end{tabular}
\caption{Examples of each different category of image augmentation, as shown on an image from the ImageNet validation set of class 259 (Pomeranian).}
\label{fig:eval-img-augs}
\end{figure}

We evaluated the robustness of each model across many different AugLy image augmentations by sampling 250 images from each dataset, computing the top-5 accuracy on those images, and computing the top-5 accuracy when the images are augmented using each augmentation. The change in top-5 accuracy from the baseline (i.e. when the images are not augmented) to the augmented images gives us a measure of how vulnerable the model is to that augmentation. We chose a diverse set of augmentations and set the parameters such that the augmentations were very noticeable but the content of the image was still recognizable to the human eye. See examples of some of the augmentations in Figure \ref{fig:eval-img-augs}. The notebook used to perform this robustness evaluation can be found in the AugLy repo at \url{https://github.com/facebookresearch/AugLy/blob/main/examples/imagenet/pwc_imagenet_v1_vs_v2_metrics.ipynb}.

\begin{figure}[h!]
\centering
\includegraphics[width=230px]{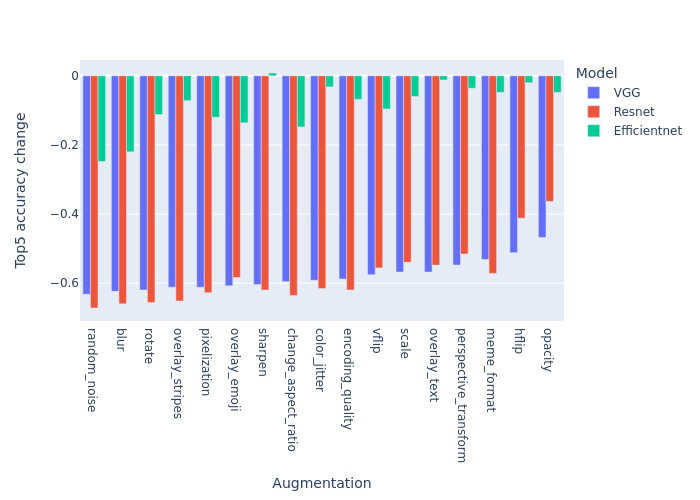}
\caption{The change in top-5 accuracy caused by each augmentation on each model, computed on a sample of 250 images from the ImageNet validation set.}
\label{fig:plot-v1}
\end{figure}

In Figure \ref{fig:plot-v1}, VGG and ResNet are pretty vulnerable to AugLy augmentations across the board. EfficientNet, on the other hand, is much more robust to most augmentations except for \texttt{blur} and \texttt{random\_noise} which cause a larger drop in accuracy. This makes sense considering the augmentations each model was trained on: VGG was trained on augmentations equivalent to AugLy's \texttt{crop}, \texttt{hflip}, \& \texttt{color\_jitter}; ResNet was trained on \texttt{crop}, \texttt{hflip}, \texttt{scale}, \& color changes similar to \texttt{color\_jitter}. EfficientNet was trained using AutoAugment\cite{autoaugment}, which includes a much wider range of augmentations such as \texttt{shear\_x/y}, \texttt{translate\_x/y}, \texttt{rotate}, \texttt{contrast}, \texttt{invert}, \texttt{solarize}, \texttt{posterize}, \texttt{color}, \texttt{brightness}, \texttt{sharpness}, and \texttt{cutout}.

Whereas VGG \& ResNet were trained on a very limited set of spatial and color-based augmentations, EfficientNet was trained on a larger number of both spatial and color-based augmentations, as well as \texttt{cutout} which is similar to the overlay augmentations in AugLy (but instead of overlaying content over the image, black rectangles are overlaid). However, none of the three models were trained on pixel-level augmentations such as \texttt{blur}, \texttt{random\_noise}, or \texttt{pixelization}, which likely explains why all three models are vulnerable to those augmentations. Figure \ref{fig:eval-img-augs} illustrates a few examples from AugLy of the four categories: spatial, color, overlay, and pixel-level augmentations.

\begin{figure}[h!]
\centering
\includegraphics[width=230px]{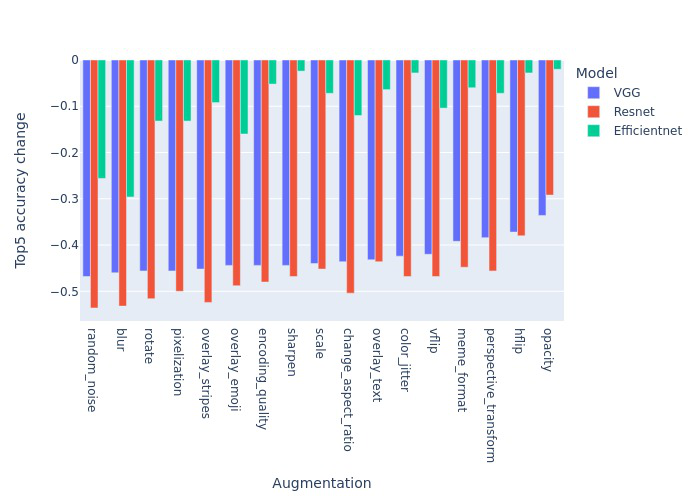}
\caption{The change in top-5 accuracy caused by each augmentation on each model, computed on a sample of 250 images from the ImageNet V2 ``matched frequency'' dataset.}
\label{fig:plot-v2}
\end{figure}

We validated that these results are comparable on the ImageNet V2 dataset, shown in Figure \ref{fig:plot-v2}. Similar to evaluation on the ImageNet validation dataset, VGG and ResNet are quite vulnerable to all augmentations at varying degrees, and EfficientNet is significantly less so with the exception of \texttt{blur} \& \texttt{random\_noise}.

\begin{figure}[h!]
\centering
\includegraphics[width=230px]{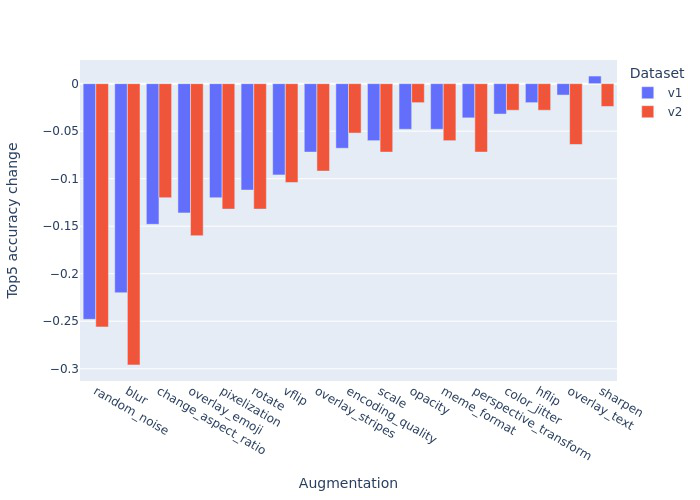}
\caption{The change in top-5 accuracy caused by each augmentation on the EfficientNet-L2 (Noisy Student) model, computed on both the ImageNet validation set \& the ImageNet V2 set.}
\label{fig:plot-eff}
\end{figure}

Figure \ref{fig:plot-eff} shows the drop in accuracy on EfficientNet for each augmentation with respect to the original ImageNet validation set and ImageNet V2. The drop in accuracy is close on both datasets for all augmentations, so there is no indication of overfitting on the validation set.

%------------------------------------------------------------------------
\section{Conclusion}

We presented AugLy, a new multimodal augmentation library with a focus on robustness. We compared each sub-library (audio, image, text, and video) to other similar augmentation libraries, assessing the amount of augmentations offered, the kinds of augmentations available, and benchmarking analogous functions to observe their performance. While other libraries may be more performant time-wise, AugLy provides a wide range of unique augmentation that replicate real modifications seen online. Additionally, we evaluated our augmentations on three state-of-the-art image classification models over time, showing that retraining on augmented data is an effective method for building defenses against various attack types.

\section*{Acknowledgements}

We would like to thank A.K.M Adib, Erik Chou, Aditya Prasad, and Guillermo Sanchez for their contributions to this paper by benchmarking and improving the efficiency of AugLy's augmentations!

{\small
\bibliographystyle{ieee_fullname}
\bibliography{egbib}

\begin{thebibliography}{10}\itemsep=-1pt

\bibitem{albumentations}
Alexander Buslaev, Alex Parinov, Eugene Khvedchenya, Vladimir~I. Iglovikov, and
  Alexandr~A. Kalinin.
\newblock Albumentations: fast and flexible image augmentations, 2020.
\newblock Information.

\bibitem{autoaugment}
Ekin~D. Cubuk, Barret Zoph, Dandelion Mane, Vijay Vasudevan, and Quoc~V. Le.
\newblock Autoaugment: Learning augmentation strategies from data, 2019.
\newblock CVPR.

\bibitem{gender_bias_dialogue}
Emily Dinan, Angela Fan, Adina Williams, Jack Urbanek, Douwe Kiela, and Jason
  Weston.
\newblock Queens are powerful too: Mitigating gender bias in dialogue
  generation, 2020.
\newblock EMNLP.

\bibitem{dfdc}
Brian Dolhansky, Joanna Bitton, Ben Pflaum, Jikuo Lu, Russ Howes, Menglin Wang,
  and Cristian~Canton Ferrer.
\newblock The deepfake detection challenge (dfdc) dataset, 2020.

\bibitem{isc}
Matthijs Douze, Giorgos Tolias, Ed Pizzi, Zoe Papakipos, Lowik Chanussot, Filip
  Radenovic, Tomas Jenicek, Maxim Maximov, Laura Leal-Taixé, Ismail Elezi,
  Ond{\v{r}}ej Chum, and Cristian~Canton Ferrer.
\newblock The 2021 image similarity challenge and dataset, 2021.

\bibitem{pytorchvideo}
Haoqi Fan, Tullie Murrell, Heng Wang, Kalyan~Vasudev Alwala, Yanghao Li, Yilei
  Li, Bo Xiong, Nikhila Ravi, Meng Li, Haichuan Yang, Jitendra Malik, Ross
  Girshick, Matt Feiszli, Aaron Adcock, Wan-Yen Lo, and Christoph
  Feichtenhofer.
\newblock {PyTorchVideo}: A deep learning library for video understanding.
\newblock In {\em Proceedings of the 29th ACM International Conference on
  Multimedia}, 2021.
\newblock \url{https://pytorchvideo.org/}.

\bibitem{nlp_data_augmentations}
Steven~Y. Feng, Varun Gangal, Jason Wei, Sarath Chandar, Soroush Vosoughi,
  Teruko Mitamura, and Eduard Hovy.
\newblock A survey of data augmentation approaches for nlp, 2021.

\bibitem{textflint}
Tao Gui, Xiao Wang, Qi Zhang, Qin Liu, Yicheng Zou, Xin Zhou, Rui Zheng, Chong
  Zhang, Jiacheng~Ye Qinzhuo~Wu, Zexiong Pang, Yongxin Zhang, Zhengyan Li,
  Ruotian Ma, Zichu Fei, Ruijian Cai, Xingwu~Hu Jun~Zhao, Zhiheng Yan, Yiding
  Tan, Yuan Hu, Qiyuan Bian, Zhihua Liu, Bolin Zhu, Shan Qin, Xiaoyu Xing,
  Jinlan Fu, Yue Zhang, Minlong Peng, Xiaoqing Zheng, Zhongyu~Wei Yaqian~Zhou,
  Xipeng Qiu, and Xuanjing Huang.
\newblock Textflint: Unified multilingual robustness evaluation toolkit for
  natural language processing, 2021.
\newblock arXiv preprint arXiv:2103.11441.

\bibitem{resnet}
Kaiming He, Xiangyu Zhang, Shaoqing Ren, and Jian Sun.
\newblock Deep residual learning for image recognition, 2015.

\bibitem{robustness_ood}
Dan Hendrycks, Steven Basart, Norman Mu, Saurav Kadavath, Frank Wang, Evan
  Dorundo, Rahul Desai, Tyler Zhu, Samyak Parajuli, Mike Guo, Dawn Song, Jacob
  Steinhardt, and Justin Gilmer.
\newblock The many faces of robustness: A critical analysis of
  out-of-distribution generalization, 2021.
\newblock ICCV.

\bibitem{robustness_perturbations}
Dan Hendrycks and Thomas Dietterich.
\newblock Benchmarking neural network robustness to common corruptions and
  perturbations, 2019.
\newblock ICLR.

\bibitem{augmix}
Dan Hendrycks, Norman Mu, Ekin~D. Cubuk, Barret Zoph, Justin Gilmer, and Balaji
  Lakshminarayanan.
\newblock Augmix: A simple data processing method to improve robustness and
  uncertainty, 2020.
\newblock ICLR.

\bibitem{audiomentations}
Iver Jordal.
\newblock Audiomentations, Aug. 2021.

\bibitem{imgaug}
Alexander~B. Jung, Kentaro Wada, Jon Crall, Satoshi Tanaka, Jake Graving,
  Christoph Reinders, Sarthak Yadav, Joy Banerjee, Gábor Vecsei, Adam Kraft,
  Zheng Rui, Jirka Borovec, Christian Vallentin, Semen Zhydenko, Kilian
  Pfeiffer, Ben Cook, Ismael Fernández, François-Michel De~Rainville,
  Chi-Hung Weng, Abner Ayala-Acevedo, Raphael Meudec, Matias Laporte, et~al.
\newblock {imgaug}.
\newblock \url{https://github.com/aleju/imgaug}, 2020.
\newblock Online; accessed 01-Feb-2020.

\bibitem{wavaugment}
Eugene Kharitonov, Morgane Rivière, Gabriel Synnaeve, Lior Wolf,
  Pierre-Emmanuel Mazaré, Matthijs Douze, and Emmanuel Dupoux.
\newblock Data augmenting contrastive learning of speech representations in the
  time domain, 2020.
\newblock CoRR, vol. abs/2007.00991.

\bibitem{vidaug}
Okan Kopuklu.
\newblock {vidaug}.
\newblock \url{https://github.com/okankop/vidaug}, 2018.

\bibitem{copy_detection}
Xiaolong Liu, Jinchao Liang, Zi-Yi Wang, Yi-Te Tsai, Chia-Chen Lin, and
  Chih-Cheng Chen.
\newblock Content-based image copy detection using convolutional neural
  network, 2020.

\bibitem{nlpaug}
Edward Ma.
\newblock {nlpaug}.
\newblock \url{https://github.com/makcedward/nlpaug}, 2019.

\bibitem{torchvision}
Francisco Massa, Vasilis Vryniotis, and Nicolas Hug.
\newblock {torchvision}.
\newblock \url{https://github.com/pytorch/vision}, 2017.

\bibitem{librosa}
Brian McFee, Colin Raffel, Dawen Liang, Daniel~P.W. Ellis, Matt McVicar, Eric
  Battenberg, and Oriol Nieto.
\newblock librosa: Audio and music signal analysis in python, 2015.
\newblock SCIPY.

\bibitem{textattack}
John~X. Morris, Eli Lifland, Jin~Yong Yoo, Jake Grigsby, Di Jin, and Yanjun Qi.
\newblock Textattack: A framework for adversarial attacks, data augmentation,
  and adversarial training in nlp.
\newblock In {\em Proceedings of the 2020 Conference on Empirical Methods in
  Natural Language Processing: System Demonstrations}, pages 119--126, 2020.

\bibitem{imagenet_v2}
Benjamin Recht, Rebecca Roelofs, Ludwig Schmidt, and Vaishaal Shankar.
\newblock Do imagenet classifiers generalize to imagenet?, 2019.

\bibitem{pydub}
James Robert.
\newblock {pydub}.
\newblock \url{https://github.com/jiaaro/pydub}, 2011.

\bibitem{data_augmentations_survey}
C. Shorten and T.M. Khoshgoftaar.
\newblock A survey on image data augmentation for deep learning, 2019.
\newblock J Big Data 6, 60.

\bibitem{vgg}
Karen Simonyan and Andrew Zisserman.
\newblock Very deep convolutional networks for large-scale image recognition,
  2015.

\bibitem{fairness_name_aug}
Eric~Michael Smith and Adina Williams.
\newblock Hi, my name is martha: Using names to measure and mitigate bias in
  generative dialogue models, 2021.

\bibitem{ssn}
Roshan Sumbaly, Mahalia Miller, Hardik Shah, Yang Xie, Sean Chang~Culatana, Tim
  Khatkevich, Enming Luo, Emanuel Strauss, Gergely Szilvasy, Manika Puri,
  Pratyusa Manadhata, Benjamin Graham, Matthijs Douze, Zeki Yalniz, and Hervé
  Jegou.
\newblock Using ai to detect covid-19 misinformation and exploitative content.
\newblock
  \url{https://ai.facebook.com/blog/using-ai-to-detect-covid-19-misinformation-and-exploitative-content/},
  2020.

\bibitem{efficientnet}
Mingxing Tan and Quoc~V. Le.
\newblock Efficientnet: Rethinking model scaling for convolutional neural
  networks, 2020.

\bibitem{torchaudio}
Yao-Yuan Yang, Moto Hira, Zhaoheng Ni, Anjali Chourdia, Artyom Astafurov,
  Caroline Chen, Ching-Feng Yeh, Christian Puhrsch, David Pollack, Dmitriy
  Genzel, Donny Greenberg, Edward~Z. Yang, Jason Lian, Jay Mahadeokar, Jeff
  Hwang, Ji Chen, Peter Goldsborough, Prabhat Roy, Sean Narenthiran, Shinji
  Watanabe, Soumith Chintala, Vincent Quenneville-Bélair, and Yangyang Shi.
\newblock Torchaudio: Building blocks for audio and speech processing.
\newblock {\em arXiv preprint arXiv:2110.15018}, 2021.

\bibitem{videomix}
Sangdoo Yun, Seong~Oh Joon, Byeongho Heo, Dongyoon Han, and Jinhyung Kim.
\newblock Videomix: Rethinking data augmentation for video classification,
  2020.

\bibitem{fairness_gender_aug}
Jieyu Zhao, Tianlu Wang, Mark Yatskar, Vicente Ordonez, and Kai-Wei Chang.
\newblock Gender bias in coreference resolution: Evaluation and debiasing
  methods.
\newblock In Proceedings of the 2018 Conference of the North American Chapter
  of the Association for Computational Linguistics: Human Language
  Technologies, Volume 2 (Short Papers), pages 15–20, New Orleans, Louisiana.
  Association for Computational Linguistics.

\bibitem{moviepy}
Zulko.
\newblock {moviepy}.
\newblock \url{https://github.com/Zulko/moviepy}, 2018.

\end{thebibliography}
}

\section*{Appendix}

\begin{figure}[h!]
\centering
\begin{tabular}{|c|c|c|c|c|}
    \hline
    \textbf{Augmentation} & \textbf{(1)} & \textbf{(2)} & \textbf{(3)} & \textbf{(4)} \\
    \hline
    Harmonic & \textbf{2.897} & & & \\
    \hline
    Percussive & \textbf{2.897} & & & \\
    \hline
    PitchShift & 1.238 & & \textbf{0.372} & 0.651 \\
    \hline
    TimeStretch & 0.415 & & \textbf{0.053} & 0.121 \\
    \hline
    Reverb & 0.271 & & \textbf{0.267} & \\
    \hline
    Tempo & \textbf{0.195} & & & \\
    \hline
    AddBackgroundNoise & 0.048 & & & \textbf{0.019} \\
    \hline
    PeakingEqualizer & \textbf{0.048} & & & \\
    \hline
    InsertInBackground & \textbf{0.044} & & & \\
    \hline
    ChangeVolume & 0.035 & \textbf{3e-5} & 0.034 & 0.004 \\
    \hline
    HighPassFilter & 0.017 & \textbf{3e-4} & 0.017 & 0.413 \\
    \hline
    ToMono & \textbf{0.016} & & 0.022 & \\
    \hline
    Normalize & 0.015 & \textbf{4e-5} & 0.043 & 0.004 \\
    \hline
    LowPassFilter & 0.014 & \textbf{5e-4} & 0.013 & 0.163 \\
    \hline
    Clicks & \textbf{0.009} & & & \\
    \hline
    Loop & \textbf{0.006} & & & \\
    \hline
    InvertChannels & \textbf{0.003} & & & \\
    \hline
    ApplyLambda & \textbf{0.003} & & & \\
    \hline
    Speed & 0.002 & \textbf{6e-5} & & \\
    \hline
    Clip & \textbf{0.002} & & & 0.003 \\
    \hline
\end{tabular} \\
\caption{The runtime (in seconds) of audio augmentations in \textbf{(1)} AugLy, \textbf{(2)} pydub, \textbf{(3)} torchaudio, \& \textbf{(4)} audiomentations.}
\label{fig:full_audio_benchmarking}
\end{figure}

\begin{figure}[h!]
\centering
\begin{tabular}{|c|c|c|c|c|}
    \hline
    \textbf{Augmentation} & \textbf{(1)} & \textbf{(2)} & \textbf{(3)} & \textbf{(4)} \\
    \hline
    ShufflePixels & \textbf{1.600} & & & \\
    \hline
    PerspectiveTransform & 0.333 & 0.032 & 0.076 & \textbf{0.013} \\
    \hline
    Sharpen & 0.159 & 0.021 & 0.141 & \textbf{0.005} \\
    \hline
    ApplyPILFilter & \textbf{0.117} & & & \\
    \hline
    ColorJitter & 0.108 & 0.038 & 0.107 & \textbf{0.015} \\
    \hline
    Blur & 0.097 & 0.013 & 0.143 & \textbf{0.005} \\
    \hline
    Saturation & 0.091 & 1.301 & 0.057 & \textbf{0.015} \\
    \hline
    ChangeAspectRatio & \textbf{0.091} & & & \\
    \hline
    Skew & \textbf{0.084} & & & \\
    \hline
    OverlayStripes & \textbf{0.083} & & & \\
    \hline
    Pixelization & 0.081 & \textbf{0.034} & & \\
    \hline
    Brightness & 0.078 & & 0.056 & \textbf{0.005} \\
    \hline
    OverlayOnto & & & & \\
    Screenshot & \textbf{0.064} & & & \\
    \hline
    Scale & \textbf{0.059} & & & \\
    \hline
    Resize & 0.056 & 0.014 & 0.050 & \textbf{0.006} \\
    \hline
    OverlayOnto & & & & \\
    BackgroundImage & \textbf{0.049} & & & \\
    \hline
    EncodingQuality & 0.041 & 0.050 & & \textbf{0.002} \\
    \hline
    ConvertColor & \textbf{0.039} & & & \\
    \hline
    MaskedComposite & \textbf{0.038} & & & \\
    \hline
    Contrast & 0.031 & \textbf{0.007} & 0.074 & \\
    \hline
    Opacity & \textbf{0.029} & & & \\
    \hline
    OverlayText & \textbf{0.027} & & & \\
    \hline
    MemeFormat & \textbf{0.025} & & & \\
    \hline
    Rotate & 0.024 & 0.019 & \textbf{0.011} & 0.028 \\
    \hline
    OverlayImage & \textbf{0.023} & & & \\
    \hline
    Pad & 0.010 & 0.018 & \textbf{0.005} & 0.008 \\
    \hline
    ApplyLambda & 0.008 & & & \textbf{2e-5} \\
    \hline
    PadSquare & \textbf{0.008} & & & \\
    \hline
    OverlayEmoji & \textbf{0.006} & & & \\
    \hline
    Grayscale & 0.005 & 0.030 & 0.002 & \textbf{0.001} \\
    \hline
    HFlip & 0.005 & 0.002 & 0.003 & \textbf{0.001} \\
    \hline
    VFlip & 0.003 & 0.001 & 0.002 & \textbf{0.001} \\
    \hline
    ClipImageSize & \textbf{0.002} & & & \\
    \hline
    Crop & 0.001 & 0.008 & 6e-4 & \textbf{2e-5} \\
    \hline
\end{tabular} \\
\caption{The runtime (in seconds) of image augmentations in \textbf{(1)} AugLy, \textbf{(2)} imgaug, \textbf{(3)} torchvision, \& \textbf{(4)} Albumentations.}
\label{fig:full_image_benchmarking}
\end{figure}

\begin{figure}[h!]
\centering
\begin{tabular}{|c|c|c|c|c|}
    \hline
    \textbf{Augmentation} & \textbf{(1)} & \textbf{(2)} & \textbf{(3)} & \textbf{(4)} \\
    \hline
    SimulateTypos & 0.276 & 0.101 & 0.006 & \textbf{4e-4} \\
    \hline
    SwapGendered & & & & \\
    Words & 0.102 & & & \textbf{0.003} \\
    \hline
    Replace & & & & \\
    FunFonts & \textbf{0.102} & & & \\
    \hline
    ReplaceSimilar & & & & \\
    UnicodeChars & \textbf{0.102} & & & \\
    \hline
    Replace & & & & \\
    UpsideDown & \textbf{0.102} & & & \\
    \hline
    MergeWords & \textbf{0.102} & & & \\
    \hline
    Replace & & & & \\
    SimilarChars & 0.102 & 0.101 & 0.006 & \textbf{0.001} \\
    \hline
    SplitWords & 0.101 & \textbf{0.101} & & \\
    \hline
    ReplaceWords & \textbf{0.101} & & & \\
    \hline
    GetBaseline & \textbf{0.101} & & & \\
    \hline
    Contractions & 0.001 & & \textbf{1e-4} & 2e-4 \\
    \hline
    ChangeCase & 4e-4 & & & \textbf{3e-4} \\
    \hline
    Insert & & & & \\
    Punctuation & & & & \\
    Chars & \textbf{1e-4} & & 0.002 & 6e-4 \\
    \hline
    Insert & & & & \\
    Whitespace & & & & \\
    Chars & \textbf{7e-5} & & & \\
    \hline
    Replace & & & & \\
    Bidirectional & \textbf{7e-5} & & & \\
    \hline
    InsertZero & & & & \\
    WidthChars & \textbf{6e-5} & & & \\
    \hline
    ApplyLambda & \textbf{6e-5} & & & \\
    \hline
\end{tabular} \\
\caption{The runtime (in seconds) of text augmentations in \textbf{(1)} AugLy, \textbf{(2)} nlpaug, \textbf{(3)} TextAttack, \& \textbf{(4)} textflint.}
\label{fig:full_text_benchmarking}
\end{figure}

\begin{figure}[h!]
\centering
\begin{tabular}{|c|c|c|c|c|}
    \hline
    \textbf{Augmentation} & \textbf{(1)} & \textbf{(2)} & \textbf{(3)} & \textbf{(4)} \\
    \hline
    ReplaceWith & & & & \\
    ColorFrames & \textbf{2.241} & & & \\
    \hline
    Loop & 2.015 & \textbf{2e-4} & & \\
    \hline
    Perspective & & & & \\
    Transform & & & & \\
    AndShake & \textbf{2.015} & & & \\
    \hline
    BlendVideos & \textbf{1.905} & & & \\
    \hline
    ReplaceWith & & & & \\
    Background & \textbf{1.746} & & & \\
    \hline
    OverlayText & \textbf{1.677} & & & \\
    \hline
    OverlayShapes & \textbf{1.659} & & & \\
    \hline
    TimeDecimate & \textbf{1.636} & & & \\
    \hline
    OverlayOnto & & & & \\
    Screenshot & \textbf{1.632} & & & \\
    \hline
    OverlayDots & \textbf{1.631} & & & \\
    \hline
    MemeFormat & \textbf{1.610} & & & \\
    \hline
    InsertIn & & & & \\
    Background & \textbf{1.605} & & & \\
    \hline
    OverlayOnto & & & & \\
    BackgroundVideo & \textbf{0.781} & & & \\
    \hline
    Shift & 0.773 & & & \textbf{0.016} \\
    \hline
    Pixelization & \textbf{0.662} & & & 1.996 \\
    \hline
    AugmentAudio & 0.625 & \textbf{0.001} & & \\
    \hline
    OverlayEmoji & \textbf{0.501} & & & \\
    \hline
    Concat & \textbf{0.467} & & & \\
    \hline
    AudioSwap & \textbf{0.445} & & & \\
    \hline
    VStack & \textbf{0.435} & & & \\
    \hline
    Overlay & \textbf{0.406} & & & \\
    \hline
    HStack & \textbf{0.400} & & & \\
    \hline
    Pad & 0.400 & \textbf{0.018} & & \\
    \hline
    TimeCrop & 0.395 & & & \textbf{1e-5} \\
    \hline
    Trim & \textbf{0.386} & & & \\
    \hline
    Change & & & & \\
    AspectRatio & \textbf{0.368} & & & \\
    \hline
    Crop & 0.352 & 9e-5 & & \textbf{2e-5} \\
    \hline
    Rotate & 0.336 & \textbf{1e-4} & 0.202 & 0.275 \\
    \hline
    Blur & 0.307 & & \textbf{0.140} & 0.179 \\
    \hline
    VFlip & 0.297 & 9e-5 & 0.151 & \textbf{2e-5} \\
    \hline
    AddNoise & 0.297 & & & \textbf{0.036} \\
    \hline
    Resize & 0.289 & \textbf{0.015} & & \\
    \hline
    Scale & \textbf{0.284} & & & \\
    \hline
    FPS & \textbf{0.271} & & & \\
    \hline
    ChangeVideo & & & & \\
    Speed & 0.269 & 1e-4 & & \textbf{1e-4} \\
    \hline
    HFlip & 0.269 & 1e-4 & 0.152 & \textbf{2e-5} \\
    \hline
    Grayscale & 0.266 & \textbf{0.047} & 0.081 & \\
    \hline
    Contrast & \textbf{0.264} & & & \\
    \hline
    Encoding & & & & \\
    Quality & \textbf{0.262} & & & \\
    \hline
    ColorJitter & 0.262 & \textbf{0.035} & 0.077 & \\
    \hline
    Brightness & 0.258 & & \textbf{0.050} & \\
    \hline
    RemoveAudio & \textbf{0.255} & & & \\
    \hline
    ApplyLambda & \textbf{5e-4} & & & \\
    \hline
\end{tabular}
\caption{The runtime (in seconds) of video augmentations in \textbf{(1)} AugLy, \textbf{(2)} moviepy, \textbf{(3)} pytorchvideo, \& \textbf{(4)} vidaug.}
\label{fig:full_video_benchmarking}
\end{figure}

\end{document}